\documentclass[12pt,a4paper]{article}
\usepackage[utf8]{inputenc}
\usepackage{amsmath}
\usepackage{amsfonts}
\usepackage{amssymb}
\usepackage{changepage}
\usepackage{bm}
\usepackage[english]{babel}
\usepackage{graphicx}
\usepackage{hyperref}
\usepackage[margin=1.0in]{geometry}
\usepackage[ruled,linesnumbered]{algorithm2e}
\usepackage{mathtools}
\usepackage{float}

\usepackage[backend=bibtex, maxcitenames=2]{biblatex}
\DeclareCiteCommand{\citeauthor} 
  {\boolfalse{citetracker}%
   \boolfalse{pagetracker}%
   \usebibmacro{prenote}}
  {\ifciteindex
     {\indexnames{labelname}}
     {}%
   \printtext[bibhyperref]{\printnames{labelname}}}
  {\multicitedelim}
  {\usebibmacro{postnote}}
\DeclareCiteCommand{\citeyear}
    {}
    {\bibhyperref{\printdate}}
    {\multicitedelim}
    {}
\newcommand{\citep}[1]{\citeauthor{#1}, \citeyear{#1}}
\newcommand{\cited}[1]{\citeauthor{#1} (\citeyear{#1})}
\newcommand{\citebracket}[1]{\cite{#1}}

\SetArgSty{textnormal}
\renewcommand{\thefootnote}{\fnsymbol{footnote}}
\graphicspath{ {./} }
\newcommand{\x}{\textbf{x}}

\newcommand{\ptheta}{p_\theta}
\newcommand{\boldmu}{\boldsymbol{\mu}}
\newcommand{\constantA}{\sqrt{\frac{1 - \alpha_{t-1}}{1-\alpha_t}}}
\newcommand{\etheta}{\epsilon_{\theta}}
\newcommand{\Qnextstep}{q(\x_{t-1} \vert \x_t, \x_0)}
\newcommand{\Pnextstep}{\ptheta(\x_{t-1} \vert \x_t)}
\newcommand{\fxt}{f_\theta(\x_t)}
\newcommand{\norm}[1]{\left\lVert#1\right\rVert}
\newcommand{\constantB}{\sqrt{\alpha_t}\x_0 + \sqrt{1-\alpha_t}\epsilon_t}
\newcommand{\pstudent}{p_{\text{student}}}
\newcommand{\pteacher}{p_{\text{teacher}}}
\newcommand{\Fteacher}{\mathcal{F}_\text{teacher}}
\newcommand{\Fstudent}{\mathcal{F}_\text{student}}
\newcommand{\stdnormal}{\mathcal{N}(\textbf{0}, \textbf{I})}

\begin{document}

\begin{center}
\begin{huge}
Knowledge Distillation in Iterative Generative Models for Improved Sampling Speed\\
\end{huge}
\end{center}

\vspace{0.25cm}

\begin{center}
\begin{large}

\textbf{Eric Luhman}\footnote{\label{note0}Equal contribution} \hspace{4cm} \textbf{Troy Luhman}\footnotemark[1] \\
ericluhman2@gmail.com \hspace{4cm} troyluhman@gmail.com
\end{large}
\end{center}
\vspace{0.25cm}

\renewcommand{\thefootnote}{\arabic{footnote}}
\setcounter{footnote}{0}



\begin{center}
\textbf{Abstract}
\\

\end{center}
\begin{changemargin}{1.75cm}{1.75cm} 
Iterative generative models, such as noise conditional score networks and denoising diffusion probabilistic models, produce high quality samples by gradually denoising an initial noise vector. However, their denoising process has many steps, making them 2-3 orders of magnitude slower than other generative models such as GANs and VAEs. In this paper, we establish a novel connection between knowledge distillation and image generation with a technique that distills a multi-step denoising process into a single step, resulting in a sampling speed similar to other single-step generative models. Our \textit{Denoising Student} generates high quality samples comparable to GANs on the CIFAR-10 and CelebA datasets, without adversarial training. We demonstrate that our method scales to higher resolutions through experiments on $256 \times 256$ LSUN. Code and checkpoints are available at \url{https://github.com/tcl9876/Denoising_Student}

\end{changemargin}


\section{Introduction}

Image Generation is an important and well studied problem in computer vision. There are a variety of approaches to image generation which yield high quality results, such as Generative Adversarial Networks (GANs, \citep{goodfellow2014generative}), Variational Autoencoders (VAEs, \citep{kingma2014autoencoding}), and energy-based models (\citep{lecun2006tutorial}). GANs generally have been able to generate the highest quality images, especially at higher resolutions (\citep{stylegan2}; \citep{stylegan2ada}; \citep{biggan}). However, their adversarial training procedure makes them more difficult to train than other methods of generative modeling. 

Two increasingly popular methods of generative modeling are Denoising score-matching (\citep{vincent2011connection}; \citep{song2019generative}) and Denoising Diffusion Probabilistic Models (DDPMs, \citep{sohldickstein2015deep}; \citep{ddpm}), both of which model data by gradually reversing a noise-adding process. Denoising score-matching methods use a neural network such as a Noise conditional score network (NCSN) to estimate the score, or gradient of the logarithmic data density.  DDPMs are trained to reverse each step of the noise-adding (a.k.a. diffusion) process, and can also be parameterized to implicitly estimate scores (\citep{ddpm}; \citep{sde}b). We refer to both models as \textit{score-based generative models}. Both approcahes use MCMC methods similar to Langevin dynamics during sampling time. Score based models are capable of producing samples that rival even the best GAN methods, without adversarial training  (\citep{ddpm}; \citep{sde}b).

A major downside to score-based generative models is that they require performing expensive MCMC sampling, often with a thousand steps or more. As a result, they can be up to three orders of magnitude slower than GANs, which only require a single network evaluation. To address this issue, Denoising Diffusion Implicit Models, or DDIMs, have been proposed (\citep{ddim}a). DDIMs use a generative Markov chain to reverse a non-Markovian inference process, and are a type of implicit probabilistic model (\citep{mohamed2017learning}). Unlike DDPMs and NCSNs, they have a \textit{deterministic} generative process that only depends on an initial latent variable. Nevertheless, DDIMs are still rather inefficient, requiring 20-100 function evaluations to produce good samples.

In this work, we propose a method of approximating a multi-step generative process with only \textit{a single function evaluation}, through the use of knowledge distillation (\citep{bucilua2006model}; \citep{hinton2015distilling}), a technique of compressing a computationally expensive teacher into a smaller student model that learns to approximate the teacher's output distribution. Our \textit{Denoising Student} synthesizes data directly from Gaussian noise without any intermediate denoising steps, and is trained to predict the output of a DDIM given the same initial noise vector. 

Our approach has a number of desirable properties. Firstly, it is far more efficient ($20 \times$ to $1000 \times$) than existing score-based approaches, making it similar to GANs and VAEs in sampling speed. Secondly, it has a simple, stable objective that does not involve any adversarial training or surrogate losses (e.g. the encoder in VAEs). Thirdly, it can be easily applied to any iterative model with a deterministic generative process, as it involves no additional architectural considerations or hyperparameters (e.g. the noise schedule, the stepsize of Langevin dynamics). Lastly, it retains the abilities of other implicit models, such as semantically meaningful interpolations through the latent space.

Despite these advantages, it produces high fidelity samples comparable to GANs on lower resolution datasets such as CIFAR-10 and CelebA $64 \times 64$. It also scales to higher resolutions, generating decent quality samples of size $256 \times 256$. To the best of our knowledge, Denoising Student is the first non-adversarial model that can produce $256 \times 256$ LSUN images in only a single step.

\section{Knowledge Distillation in Deterministic Generative Models}

\subsection{Knowledge Distillation}
Knowledge distillation (\citep{bucilua2006model}; \citep{hinton2015distilling}) involves compressing an expensive but high-performing teacher model into a smaller student model. The student model is trained to minimize the cross-entropy between its output distribution and the output distribution of the teacher, which results in better performance than if it was trained normally. In supervised learning tasks, the motivation for this approach comes from the information hidden in the teacher's output, which isn't present in a one-hot label. For instance, an image with two objects could be plausibly classified as either one, which is reflected in the teacher's output distribution, but not the original label. 

One condition of knowledge distillation is that the function to be learned must be deterministic, otherwise, it will be difficult, if not impossible to learn. In supervised learning tasks, this is a rather trivial condition, since models produce the same output given the same input. In generative modeling, however, this condition is no longer trivial. Score-based and energy-based models use a stochastic MCMC procedure such as Langevin dynamics, making their usage as a teacher infeasible. We could apply knowledge distillation to a GAN, which is deterministic given the same latent variable, but we would gain only a minor speedup over an already fast model.  

Recently, \citeauthor{ddim} (\citeyear{ddim}a) proposed a new class of generative models called \textit{denoising diffusion implicit models} (DDIMs), which are similar to other iterative generative models in terms of sample quality but have somewhat faster sampling speed. Unlike other iterative models\footnote{An even more recent work (\citep{sde}b) proposed a neural \textit{probability flow ODE} that also has a deterministic, multi-step generative process. However, pretrained models are unavailable as of writing, and we wanted to avoid retraining the teacher.}, their generative process is deterministic, which motivates their usage as a teacher model. We will discuss them below. For further explanation and proofs related to DDIMs, the reader is referred to \citeauthor{ddim} (\citeyear{ddim}a).

\subsection{Denoising Diffusion Implicit Models}

Consider an inference process with latent variables $\x_1, \ldots, \x_T$ of the same dimensionality as the data $\x_0 \sim q(\x_0)$, parameterized by a decreasing sequence of constants $\alpha_{1:T} \in (0,1]^T$. The inference process $q(\x_{1:T} \vert \x_0)$ is defined as:

\begin{gather}
q(\x_{1:T} \vert \x_0) \coloneqq q(\x_T \vert \x_0) \prod_{t=2}^{T} \Qnextstep, \text{ with} \\
q(\x_T \vert \x_0) = \mathcal{N}(\x_T; \sqrt{\alpha_T}\x_0, (1 - \alpha_T)\textbf{I}), \text{ and } \\[10pt]
\Qnextstep = \delta(\x_{t-1} - \boldmu(\x_t, \x_0))
\end{gather}

where $\delta(\cdot)$ is the Dirac delta function, $\boldmu(\x_t, \x_0) = \sqrt{\alpha_{t-1}}\x_0 + \constantA (\x_t - \sqrt{\alpha_t}\x_0)$. This inference process is not trainable, unlike in other probabilistic models such as VAEs (\citep{kingma2014autoencoding}). One notable property of this inference process is that for any $t$, $q(\x_t \vert \x_0) = \mathcal{N}(\x_t; \sqrt{\alpha_t} \x_0, (1-\alpha_t)\textbf{I})$, so we can write:
\begin{equation}\label{eq4}
\x_t = \sqrt{\alpha_t} \x_0 + \sqrt{1-\alpha_t} \epsilon, \epsilon \sim \stdnormal
\end{equation}
A denoising diffusion implicit model (\citep{ddim}a) is a latent variable model whose generative process reverses the above inference process. Formally, it is defined as:
\begin{equation}\label{eq5}
\ptheta(\x_0) \coloneqq \int \ptheta(\x_{0:T}) \text{d} \x_{1:T}, \text{ with }
\ptheta(\x_{0:T}) \coloneqq p(\x_T) \prod_{t=1}^{T} \Pnextstep
\end{equation}
where $\ptheta(\x_{0:T})$ is the generative process, with prior $p(\x_T) = \mathcal{N}(\x_T; \textbf{0}, \textbf{I})$. 
Each distribution $\Pnextstep$ in the generative process estimates the corresponding inference distribution $\Qnextstep$. However, sampling from $\Qnextstep$ requires knowledge of $\x_0$, which is unavailable during the generative process. As a result, we make a prediction $\fxt$ of $\x_0$, and use this in our estimate $q(\x_{t-1} \vert \x_t, \fxt)$. 
Putting it together, we get
\begin{align}\label{eq6}
\Pnextstep =
 \begin{cases}
 q(\x_{t-1} \vert \x_t, \fxt) & \text{if $t>1$} \\
 \delta(\x_0 - f_\theta(\x_1)) & \text{if $t=1$} 
 \end{cases}
\end{align} 
where $\fxt$ is found by rewriting Eq. \eqref{eq4}, to get $\fxt = \frac{1}{\sqrt{\alpha_t}} (\x_t - \sqrt{1 - \alpha_t} \etheta)$. To sample from a DDIM, we first sample a noise vector $\x_T$ from a standard normal, then gradually denoise it over time by iteratively computing  $\x_{T-1}, \x_{T-2}, \ldots, \x_1, \x_0$ with:
\begin{equation}\label{eq7}
\sqrt{\frac{1}{\alpha_{t-1}}} \x_{t-1} = \sqrt{\frac{1}{\alpha_t}} \x_t - \left(\sqrt{\frac{1-\alpha_{t}}{\alpha_{t}}} - \sqrt{\frac{1-\alpha_{t-1}}{\alpha_{t-1}}} \right) \etheta(\x_t) , \text{ or }
\end{equation}
\begin{equation}\label{eq8}
\x_{t-1} = \sqrt{\frac{\alpha_{t-1}}{\alpha_t}}(\x_t - \sqrt{1-\alpha_t}\etheta(\x_t)) + \sqrt{1-\alpha_{t-1}}\etheta(\x_t)
\end{equation}
The generative model is trained by maximizing the variational lower bound: \\ $\text{ELBO} = \mathbb{E}_q [\log p_\theta(\x_{0:T}) - \log q(\x_{1:T} \vert \x_0)]$. Here, we treat $\Pnextstep$ and each $\Qnextstep$ as isotropic Gaussians with very small variance, to ensure that the generative and inference processes are nonzero everywhere.\footnote{Their original definitions were chosen to emphasize their non-stochasticity.} This objective can then be simplified to a closed form expression:
\begin{equation}\label{eq9}
L = \sum_{t=1}^{T} \gamma_t \mathbb{E}_{\x_0, \epsilon_t} \left[ \norm{\epsilon_t - \etheta(\constantB, t)}_2^2 \right] + C,
\end{equation}
where $\gamma_t$ is a positive constant\footnote{In practice, \cited{ddpm} found it beneficial to set $\gamma_t = 1$ for all $t$} related to $\alpha_t$ and $\alpha_{t-1}$, and $\etheta$ is a trainable function (i.e. neural network) that approximates $\epsilon_t$ given a noisy input $\x_t$ and timestep $t$. This is the same training objective used by DDPMs (\citep{ddpm}), so one can use a trained DDPM as a DDIM by changing only the sampling procedure.

From Eq. \eqref{eq8}, we see that for a given $\etheta$, $\x_{t-1}$ is known and fixed given $\x_t$. In other words, each individual step of the sampling procedure is deterministic. As a result, the entire generative process from $\x_{T-1}$ to $\x_0$ is deterministic as well, dependent only on the initial latent $\x_T$. So, one could choose to ignore the intermediate latents $\x_{1:T-1}$, since knowledge of them is not necessary to predict $\x_0$. We can therefore directly model the data $\x_0$ with a conditional distribution $p_\theta(\x_0 \vert \x_T)$. 

\begin{figure}
\includegraphics[width=15.0cm, height=4.54cm]{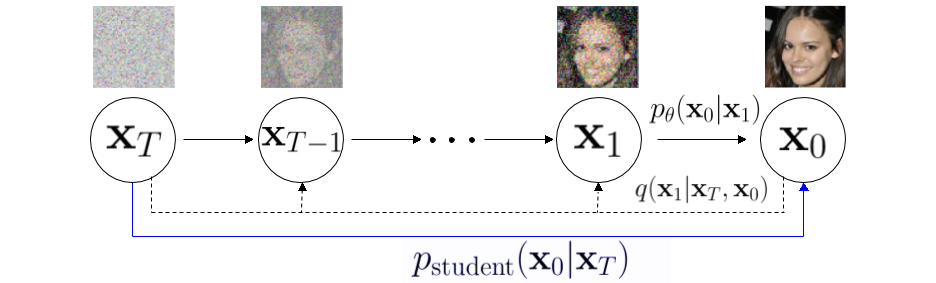}
\caption{A graphical representation of our method.}
\label{fig0graphic}
\end{figure}

From this perspective, we could think of a DDIM as an implicit probabilistic model that transforms a latent variable $\x_T$ with a deterministic function to model data. Unfortunately, evaluating this function requires \textit{T} forward passes through a neural network, making it very computationally expensive. This motivates our usage of knowledge distillation, where we distill the knowledge in this ``teacher'' function into a much faster ``student'' network that uses only a single network evaluation. The student's output distribution $\pstudent(\x_0 \vert \x_T)$ approximates the teacher's output distribution $\pteacher(\x_0 \vert \x_T)$, using only $\x_T$. We illustrate this concept in Figure \ref{fig0graphic}.

\section{Generative Modeling with Denoising Student} 

\subsection{Training Objective}
As previously mentioned, the student's conditional distribution $\pstudent(\x_0 \vert \x_T)$ should be as close as possible close to the teacher's conditional distribution $\pteacher(\x_0 \vert \x_T)$. We therefore train the student by minimizing the following:

\begin{equation}\label{eq10}
L_\text{student} = \mathbb{E}_{\x_T} \left[D_{\text{KL}}( \pteacher(\x_0 \vert \x_T) \ \Vert \ \pstudent(\x_0 \vert \x_T)) \right]
\end{equation}
This training objective is very similar to that in a supervised learning task, where models are trained minimize the KL Divergence between two categorical distributions over labels. 

We parameterize the student as a simple Gaussian model with trainable mean $\Fstudent(\x_T)$ and unit variance. Since the teacher is a deterministic function $\Fteacher$ of $\x_T$, we add some Gaussian noise to its output to ensure that $\pteacher(\x_0 \vert \x_T)$ is nonzero everywhere.  Since the KL term in Eq. \eqref{eq10} is between two Gaussians, the equation simplifies to:
\begin{equation}\label{eq11}
L_\text{student} = \frac{1}{2} \mathbb{E}_{\x_T} [\norm{\Fstudent(\x_T) - \Fteacher(\x_T)}_{2}^{2}]+C
\end{equation}
Training of the student is done by sampling random $\x_T$ from the prior $p(\x_T)$ and computing the corresponding $\Fteacher(\x_T)$ with a DDIM, then minimizing Eq. \eqref{eq11}. Unlike other generative models, our method does not require the joint training of two networks (e.g. discriminator in GANs, inference model in VAEs). In theory, one could train the student on as many unique examples as desired, since both the prior and the teacher model are known. In our experiments however, we use only 1.024 million synthetic examples for each dataset and iterate over these same ones multiple times.

\subsection{Model Architecture}
Unlike normalizing flows and autoregressive models, our method does not have any significant architectural restrictions. So, one could use an arbitrary neural network as the student, then randomly initialize weights.
But we observe that the entirety of the teacher's knowledge is contained in a much smaller neural network $\etheta$ that is reused many times. As such, we initialize the student to have \textit{the same architecture and weights} as $\etheta$. This allows the student to inherit knowledge from the teacher, which speeds up training. 

In the remaining sections, we use the term ``teacher'' to refer to the DDIM's sampling procedure from $\x_T$ to $\x_0$, and the term ``teacher network'' to refer to the neural network $\etheta(\x_t, t)$ that parameterizes this sampling procedure.

There are a couple minor concerns with this approach that we will address here. Firstly, the student is trained to model the data $\x_0$, so it may not be ideal to initialize it to a teacher network that models the noise $\epsilon_t$. To resolve this, we have the student predict the noise as well, and subtract the predicted noise from the input noise $\x_T$ to produce samples and compute loss. Secondly, the teacher network (and therefore the student as well) is conditioned on time, so we condition the student on timestep \textit{T}, corresponding to the highest noise level.

\section{Experiments}
In this section, we examine the proposed method through multiple experiments. We consider four datasets: CIFAR-10 $32 \times 32$ \citebracket{cifar}, CelebA $64 \times 64$ \citebracket{celeba}, LSUN \citebracket{lsun} Bedroom $256 \times 256$, and LSUN Church $256 \times 256$. We use the pretrained models from \cited{ddpm} as our teacher network, except for CelebA, where we use the one from \citeauthor{ddim} (\citeyear{ddim}a). For our experiments, we use a 100-step DDIM as our teacher for CIFAR-10 and CelebA, and a 50-step DDIM for LSUN. Further details can be found in \autoref{experimentaldetails}.

\subsection{Image Generation}
We show random samples from our CIFAR-10 and CelebA model in Figures \ref{fig1cifar_samples} and \ref{fig2celeba_samples} respectively. The samples are both high quality and diverse, demonstrating the efficacy of our method. Note that unlike VAEs, we do not reduce the temperature (standard deviation) of our prior when showing qualitative samples, which hinders diversity. 

For quantitative evaluation, we show FID \citebracket{heusel2018gans} and Inception scores \citebracket{salimans2016improved} in Table \ref{tab:table1} for various methods on CIFAR-10, as well as the number of steps (network evaluations) each needs. On CIFAR-10, our model achieves an FID of 9.36, which is lower than several GANs and far lower than NVAE \citebracket{nvae}, a state of the art VAE.\footnote{While \cited{vdvae} has slightly lower negative log-likelihood than NVAE (2.87 vs 2.91), we do not believe that this improvement would fully account for the disparity between our model and NVAE.} Our CIFAR-10 scores are also similar to state-of-the-art EBMs, even though our method uses only a single network evaluation. On CelebA, we obtain a competitive FID of 10.68. 

\begin{table}[t!]
  \parbox{.6\linewidth}{
  	\begin{small}
    \caption{Results for CIFAR-10. "Steps" refers to the number of neural network evaluations needed to generate a sample, and a model labeled with "cond." is class-conditional. Scores marked with an asterisk were computed by us and are not official.} 
    \label{tab:table1}
    \begin{tabular}{l c c c} 
      \\
      Model & FID $\downarrow$ & IS $\uparrow$ & Steps $\downarrow$\\
      \hline \\
      \textbf{Single-Step} \\
      Denoising Student (\textbf{Ours}) & 9.36 & 8.36 & 1 \\
      NVAE \citebracket{nvae} & 51.67 & 5.51 & 1\\
      MoLM \citebracket{molm} & 18.9 & 7.90 & 1 \\
      \\ 
      \textbf{Single-Step, GAN} \\
      SNGAN \citebracket{sngan} & 21.7 & 8.22 & 1 \\
      BigGAN (cond.) \citebracket{biggan} & 14.73 & 9.22 & 1  \\
      PPOGAN \citebracket{ppogan} & 10.87 & 8.69 & 1 \\
      StyleGAN2+ADA \citebracket{stylegan2ada} & 2.92 & 9.83 & 1 \\
      StyleGAN2+ADA (cond.) \citebracket{stylegan2ada} & 2.42 & \textbf{10.14} & 1  \\
      \\
      \textbf{Many-Step} \\ 
      DDIM (100 step, \textbf{Teacher}) & 4.16 & 8.96* & 100 \\
      EBM \citebracket{du2019implicit} & 38.2 & 6.78 & 60 \\
      VAEBM \citebracket{vaebm} & 12.19 & 8.43 & 16  \\
      EBM+recovery likelihood \citebracket{gao2020learning} & 9.60 & 8.58 & 180  \\
      NCSNv2 \citebracket{ncsn2} & 10.87 & 8.40 & 1160  \\
      DDPM \citebracket{ddpm} & 3.17 & 9.46 & 1000  \\
      NCSN++ (8 blocks/res) \citebracket{sde} & \textbf{2.20} & 9.89 & 2000
    \end{tabular}
    \end{small}
  }
  \hfill
  \parbox{.3\linewidth}{
  	\begin{small}
    \caption{Time (s) to generate 50k CIFAR-10 images} 
    \label{tab:table2}
    \begin{tabular}{l|c} 
      \\
      Model & Time\\
      \hline \\
      Denoising Student & \textbf{51.5} \\
      NVAE  & 146.5 \\
      DDIM (Teacher)  & $4.96 \times 10^3$ \\
      DDPM & $5.02 \times 10^4$
    \end{tabular}
    \end{small}
  }
\end{table}

To see how our method scales to higher resolutions, we evaluated it on the $256 \times 256$ LSUN Bedroom and Church datasets. We show random samples in Figure \ref{fig3lsun_bedroom_samples} (Bedroom) and \ref{fig4lsun_church_samples} (Church). We find that our model learns structure and color quite well, as well as larger details such as lamps and windows in Bedroom, and clouds and trees in Church. We do find, however, that our model produces rather blurry samples, without very defined textures.  We believe this bluriness arises from the choice of objective. While GANs are designed to produce images that fool the discriminator, our model has to replicate the teacher's output down to the pixel, which is difficult for higher dimensional data.

\vspace{-0.25cm}
\begin{minipage}{0.45\linewidth} 
\begin{figure}[H]
\includegraphics[height=4.5cm, width=6.0cm]{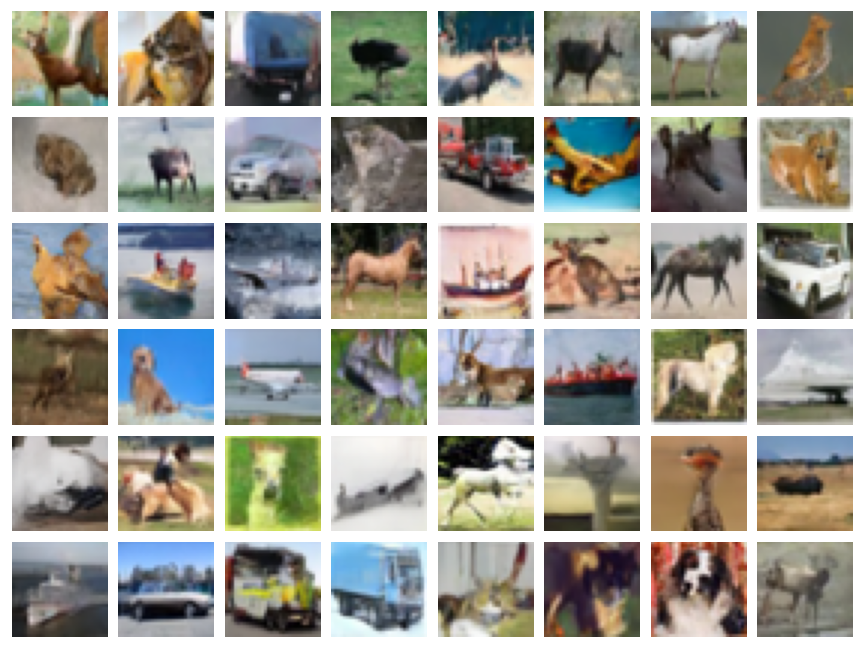}
\caption{Uncurated samples from our CIFAR-10 model}
\label{fig1cifar_samples}
\end{figure}
\end{minipage}
\hspace{1cm}
\begin{minipage}{0.45\linewidth} 
\begin{figure}[H]
\includegraphics[height=4.5cm, width=5.63cm]{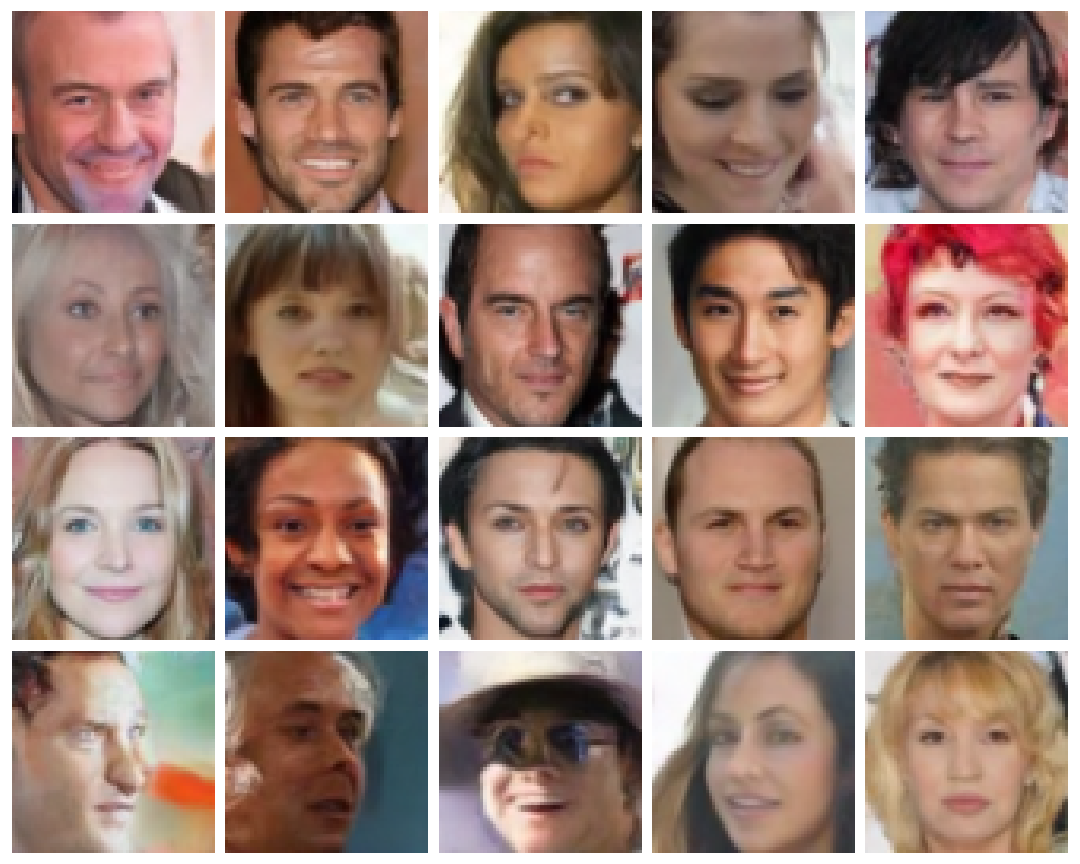}
\caption{Uncurated samples from our CelebA model}
\label{fig2celeba_samples}
\end{figure}
\end{minipage}

\vspace{-0.25cm}
\begin{minipage}{0.45\linewidth} 
\begin{figure}[H]
\includegraphics[height=5cm, width=7.5cm]{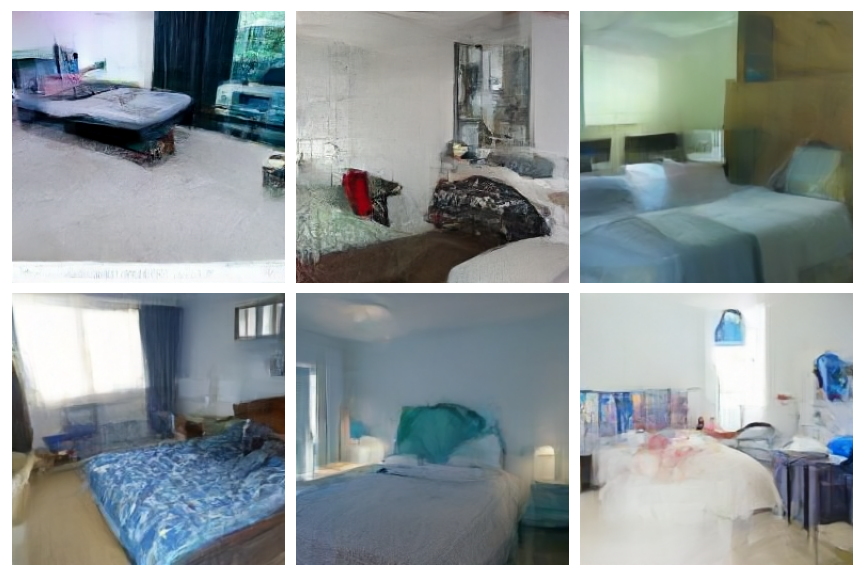}
\caption{Uncurated samples from our LSUN Bedroom model}
\label{fig3lsun_bedroom_samples}
\end{figure}
\end{minipage}
\hspace{1cm}
\begin{minipage}{0.45\linewidth} 
\begin{figure}[H]
\includegraphics[height=5cm, width=7.5cm]{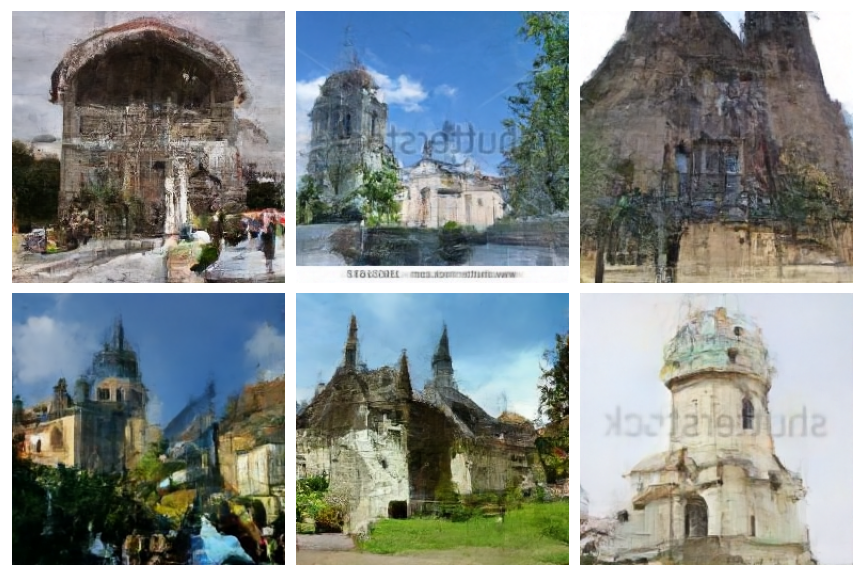}
\caption{Uncurated samples from our LSUN Church-Outdoor model}
\label{fig4lsun_church_samples}
\end{figure}
\end{minipage}

To assess the sampling speed of Denoising Student, we measured the time it took to generate 50k CIFAR-10 images, and compared it in Table \ref{tab:table2}. Notably, it is about $100 \times$ faster than the teacher, and $1000 \times$ faster than a DDPM, indicating that for the same architecture, sampling time is proportional to number of steps. When compared with NVAE, we find ours is slightly faster, showing that our model is indeed similar to other single-step models in speed. Timing experiments were done with a batch size of 250 on a single Nvidia V100 GPU. 

\subsection{Latent Space Manipulation}

\begin{figure}[H] 
\centering
\includegraphics[height=7cm, width=15.4cm]{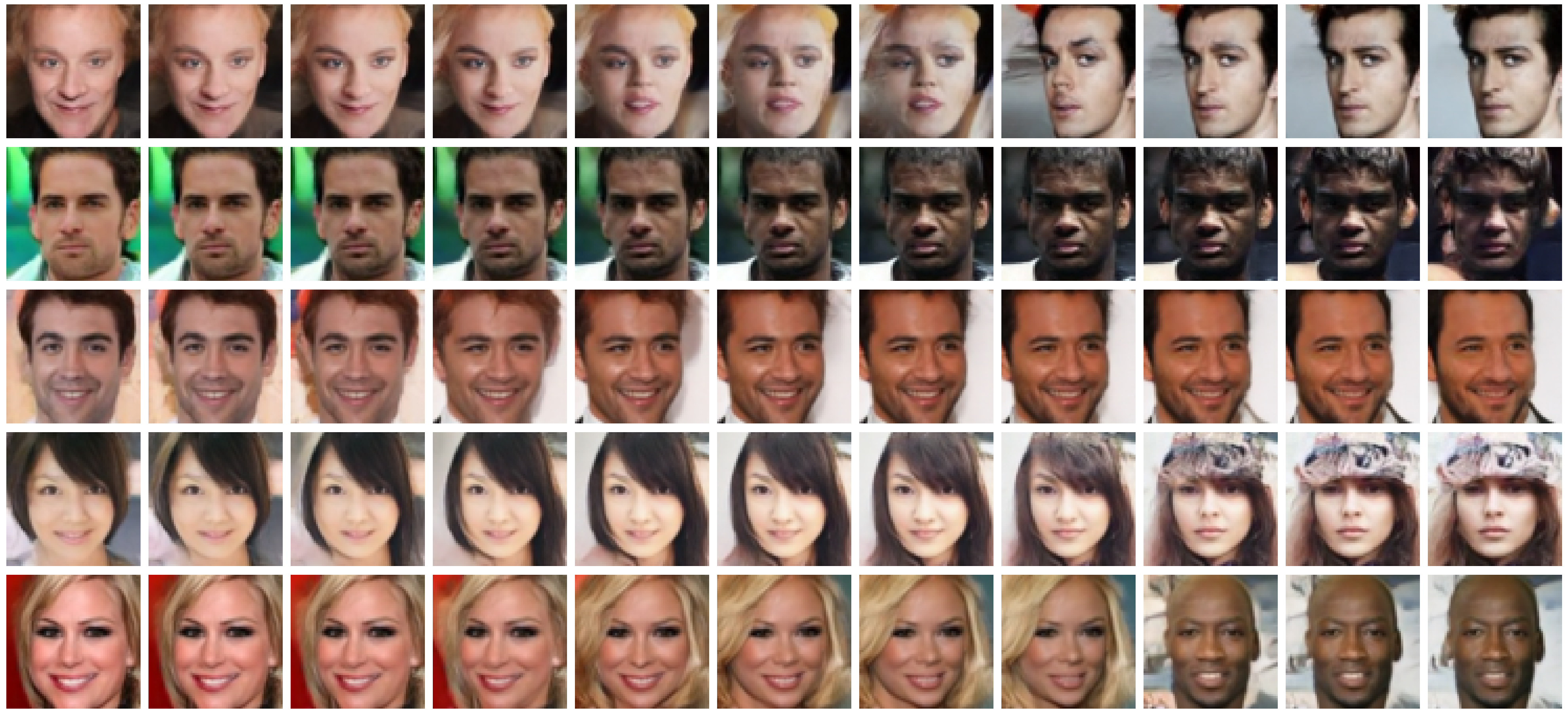}
\caption{Interpolation results on the CelebA dataset.}
\label{fig5celeba_interpolation}
\end{figure}

\citeauthor{ddim} (\citeyear{ddim}a) observed that in DDIMs, the latent variable $\x_T$ is an effective representation of the data, where interpolations in the latent space result in meaningful image interpolations. Our model was capable of learning the latent mapping of a DDIM, so we hypothesize it would be able to produce meaningful interpolations as well. To test this hypothesis, we perform spherical interpolation between 2 random $\x_T$, and show the result in Figure \ref{fig5celeba_interpolation}. See Figure \ref{fig11interpolation_results} for more.

To gain some further insight into the learned latent representation of Denoising Student, we decided to apply it to \textit{different sized} images than it was trained on, by simply changing the size of the latent variable. Remarkably, it still manages to replicate the teacher quite well, producing coherent images of a different size (Figure \ref{fig12different_resolution} in Appendix). This would indicate that the student retains the teacher's advanced generalization capabilities.

\section{Related Work}
Knowledge distillation (\citep{bucilua2006model}; \citep{hinton2015distilling}) has been explored extensively in supervised learning tasks such as image classification \citebracket{romero2015fitnets, zagoruyko2016paying, yim2017gift, furlanello2018born, tian2019contrastive, xu2020knowledge}, language modeling \citebracket{sun2019patient, turc2019wellread, sanh2020distilbert, jiao2019tinybert, sun2020contrastive}, and speech recognition \citebracket{chebotar2016distilling, Fukuda2017EfficientKD, watanabe2017student, kim2019knowledge, gao2020distilling}. Most existing work on knowledge distillation is concerned with finding the best way to transfer knowledge (e.g. best loss function) from the teacher to the student. Additionally, knowledge distillation is usually concerned with reducing the size (parameters) of a network, instead of the number of evaluations.

Our work is also based on denoising score-based models such as NCSNs (\citep{song2019generative}) and DDPMs (\citep{ddpm}), which estimate the score of the data density and use a denoising auto-encoder objective. \citeauthor{sde} (\citeyear{sde}b) observed that an NCSN / DDPM's generative process could be seen as solving a reverse time stochastic differential equation that reverses a noise-adding SDE. A DDIM's sampling procedure, on the other hand, more closely resembles integrating an ODE (see. Eq. \ref{eq7}). From this perspective, our student model could loosely be seen as approximating this integral with a function, instead of a more accurate but expensive numerical method (the teacher).

\section{Conclusion}
In this work, we presented a simple method to vastly reduce the sampling time of certain iterative generative models, with only a minor degradation in performance. Our knowledge distillation technique can be easily applied to any trained model with a deterministic generative process, since it has a stable training objective and requires no extra architectural considerations. Despite its simplicity and stability, our model is capable of producing significantly better samples than other non-adversarial, single-step methods such as VAEs. It also learns meaningful latent representations, allowing for easy data manipulation through the latent space. Future work includes bridging the gap between the teacher and student through more advanced distillation techniques, and producing sharper images at high resolutions.

\pagebreak

\printbibliography

@misc{ddpm,
      title={Denoising Diffusion Probabilistic Models}, 
      author={Jonathan Ho and Ajay Jain and Pieter Abbeel},
      year={2020},
      eprint={2006.11239},
      archivePrefix={arXiv},
      primaryClass={cs.LG}
}

@inproceedings{song2019generative,
	title={Generative modeling by estimating gradients of the data distribution},
	author={Song, Yang and Ermon, Stefano},
	booktitle={Advances in Neural Information Processing Systems},
	pages={11918--11930},
	year={2019}
}

@article{ppogan,
  title={Improving GAN Training with Probability Ratio Clipping and Sample Reweighting},
  author={Wu, Yue and Zhou, Pan and Wilson, Andrew Gordon and Xing, Eric and Hu, Zhiting},
  journal={Advances in Neural Information Processing Systems},
  volume={33},
  year={2020}
}

@misc{sngan,
      title={Spectral Normalization for Generative Adversarial Networks}, 
      author={Takeru Miyato and Toshiki Kataoka and Masanori Koyama and Yuichi Yoshida},
      year={2018},
      eprint={1802.05957},
      archivePrefix={arXiv},
      primaryClass={cs.LG}
}

@misc{ncsn2,
      title={Improved Techniques for Training Score-Based Generative Models}, 
      author={Yang Song and Stefano Ermon},
      year={2020},
      eprint={2006.09011},
      archivePrefix={arXiv},
      primaryClass={cs.LG}
}

@misc{mohamed2017learning,
      title={Learning in Implicit Generative Models}, 
      author={Shakir Mohamed and Balaji Lakshminarayanan},
      year={2017},
      eprint={1610.03483},
      archivePrefix={arXiv},
      primaryClass={stat.ML}
}

@inbook{lecun2006tutorial,
title = "A tutorial on energy-based learning",
author = "Yann Lecun and Sumit Chopra and Raia Hadsell and Ranzato, {Marc Aurelio} and Huang, {Fu Jie}",
year = "2006",
language = "English (US)",
editor = "G. Bakir and T. Hofman and B. Scholkopt and A. Smola and B. Taskar",
booktitle = "Predicting structured data",
publisher = "MIT Press",

}

@misc{adam,
      title={Adam: A Method for Stochastic Optimization}, 
      author={Diederik P. Kingma and Jimmy Ba},
      year={2017},
      eprint={1412.6980},
      archivePrefix={arXiv},
      primaryClass={cs.LG}
}

@misc{ddim,
      title={Denoising Diffusion Implicit Models}, 
      author={Jiaming Song and Chenlin Meng and Stefano Ermon},
      year={2020},
      eprint={2010.02502},
      archivePrefix={arXiv},
      primaryClass={cs.LG}
}

@misc{romero2015fitnets,
      title={FitNets: Hints for Thin Deep Nets}, 
      author={Adriana Romero and Nicolas Ballas and Samira Ebrahimi Kahou and Antoine Chassang and Carlo Gatta and Yoshua Bengio},
      year={2015},
      eprint={1412.6550},
      archivePrefix={arXiv},
      primaryClass={cs.LG}
}

@misc{heusel2018gans,
      title={GANs Trained by a Two Time-Scale Update Rule Converge to a Local Nash Equilibrium}, 
      author={Martin Heusel and Hubert Ramsauer and Thomas Unterthiner and Bernhard Nessler and Sepp Hochreiter},
      year={2018},
      eprint={1706.08500},
      archivePrefix={arXiv},
      primaryClass={cs.LG}
}

@misc{salimans2016improved,
      title={Improved Techniques for Training GANs}, 
      author={Tim Salimans and Ian Goodfellow and Wojciech Zaremba and Vicki Cheung and Alec Radford and Xi Chen},
      year={2016},
      eprint={1606.03498},
      archivePrefix={arXiv},
      primaryClass={cs.LG}
}

@article{vincent2011connection,
  title={A connection between score matching and denoising autoencoders},
  author={Vincent, Pascal},
  journal={Neural computation},
  volume={23},
  number={7},
  pages={1661--1674},
  year={2011},
  publisher={MIT Press}
}

@inproceedings{goodfellow2014generative,
 author = {Goodfellow, Ian and Pouget-Abadie, Jean and Mirza, Mehdi and Xu, Bing and Warde-Farley, David and Ozair, Sherjil and Courville, Aaron and Bengio, Yoshua},
 booktitle = {Advances in Neural Information Processing Systems},
 editor = {Z. Ghahramani and M. Welling and C. Cortes and N. Lawrence and K. Q. Weinberger},
 pages = {2672--2680},
 publisher = {Curran Associates, Inc.},
 title = {Generative Adversarial Nets},
 url = {https://proceedings.neurips.cc/paper/2014/file/5ca3e9b122f61f8f06494c97b1afccf3-Paper.pdf},
 volume = {27},
 year = {2014}
}

@misc{sohldickstein2015deep,
      title={Deep Unsupervised Learning using Nonequilibrium Thermodynamics}, 
      author={Jascha Sohl-Dickstein and Eric A. Weiss and Niru Maheswaranathan and Surya Ganguli},
      year={2015},
      eprint={1503.03585},
      archivePrefix={arXiv},
      primaryClass={cs.LG}
}

@misc{lsun,
      title={LSUN: Construction of a Large-scale Image Dataset using Deep Learning with Humans in the Loop}, 
      author={Fisher Yu and Ari Seff and Yinda Zhang and Shuran Song and Thomas Funkhouser and Jianxiong Xiao},
      year={2016},
      eprint={1506.03365},
      archivePrefix={arXiv},
      primaryClass={cs.CV}
}

@inproceedings{celeba,
  title={Deep learning face attributes in the wild},
  author={Liu, Ziwei and Luo, Ping and Wang, Xiaogang and Tang, Xiaoou},
  booktitle={Proceedings of the IEEE international conference on computer vision},
  pages={3730--3738},
  year={2015}
}

@article{cifar,
author = {Krizhevsky, Alex},
year = {2012},
month = {05},
pages = {},
title = {Learning Multiple Layers of Features from Tiny Images},
journal = {University of Toronto}
}

@misc{kingma2014autoencoding,
      title={Auto-Encoding Variational Bayes}, 
      author={Diederik P Kingma and Max Welling},
      year={2014},
      eprint={1312.6114},
      archivePrefix={arXiv},
      primaryClass={stat.ML}
}

@misc{sde,
      title={Score-Based Generative Modeling through Stochastic Differential Equations}, 
      author={Yang Song and Jascha Sohl-Dickstein and Diederik P. Kingma and Abhishek Kumar and Stefano Ermon and Ben Poole},
      year={2020},
      eprint={2011.13456},
      archivePrefix={arXiv},
      primaryClass={cs.LG}
}

@inproceedings{biggan,
title={Large Scale {GAN} Training for High Fidelity Natural Image Synthesis},
author={Andrew Brock and Jeff Donahue and Karen Simonyan},
booktitle={International Conference on Learning Representations},
year={2019},
url={https://openreview.net/forum?id=B1xsqj09Fm},
}

@inproceedings{xu2020knowledge,
  title={Knowledge distillation meets self-supervision},
  author={Xu, Guodong and Liu, Ziwei and Li, Xiaoxiao and Loy, Chen Change},
  booktitle={European Conference on Computer Vision},
  pages={588--604},
  year={2020},
  organization={Springer}
}

@inproceedings{yim2017gift,
  title={A gift from knowledge distillation: Fast optimization, network minimization and transfer learning},
  author={Yim, Junho and Joo, Donggyu and Bae, Jihoon and Kim, Junmo},
  booktitle={Proceedings of the IEEE Conference on Computer Vision and Pattern Recognition},
  pages={4133--4141},
  year={2017}
}

@inproceedings{tian2019contrastive,
  title={Contrastive Representation Distillation},
  author={Tian, Yonglong and Krishnan, Dilip and Isola, Phillip},
  booktitle={International Conference on Learning Representations},
  year={2019}
}

@inproceedings{zagoruyko2016paying,
title={Paying More Attention to Attention: Improving the Performance of Convolutional Neural Networks via Attention Transfer},
author={Zagoruyko, Sergey and Komodakis, Nikos},
booktitle={International Conference on Learning Representations},
year={2017}
}

@misc{stylegan2,
      title={Analyzing and Improving the Image Quality of StyleGAN}, 
      author={Tero Karras and Samuli Laine and Miika Aittala and Janne Hellsten and Jaakko Lehtinen and Timo Aila},
      year={2020},
      eprint={1912.04958},
      archivePrefix={arXiv},
      primaryClass={cs.CV}
}

@misc{stylegan2ada,
      title={Training Generative Adversarial Networks with Limited Data}, 
      author={Tero Karras and Miika Aittala and Janne Hellsten and Samuli Laine and Jaakko Lehtinen and Timo Aila},
      year={2020},
      eprint={2006.06676},
      archivePrefix={arXiv},
      primaryClass={cs.CV}
}

@misc{nvae,
      title={NVAE: A Deep Hierarchical Variational Autoencoder}, 
      author={Arash Vahdat and Jan Kautz},
      year={2020},
      eprint={2007.03898},
      archivePrefix={arXiv},
      primaryClass={stat.ML}
}

@misc{vaebm,
      title={VAEBM: A Symbiosis between Variational Autoencoders and Energy-based Models}, 
      author={Zhisheng Xiao and Karsten Kreis and Jan Kautz and Arash Vahdat},
      year={2020},
      eprint={2010.00654},
      archivePrefix={arXiv},
      primaryClass={cs.LG}
}

@misc{vdvae,
      title={Very Deep VAEs Generalize Autoregressive Models and Can Outperform Them on Images}, 
      author={Rewon Child},
      year={2020},
      eprint={2011.10650},
      archivePrefix={arXiv},
      primaryClass={cs.LG}
}

@misc{gao2020learning,
      title={Learning Energy-Based Models by Diffusion Recovery Likelihood}, 
      author={Ruiqi Gao and Yang Song and Ben Poole and Ying Nian Wu and Diederik P. Kingma},
      year={2020},
      eprint={2012.08125},
      archivePrefix={arXiv},
      primaryClass={cs.LG}
}

@misc{molm,
      title={Learning Implicit Generative Models with the Method of Learned Moments}, 
      author={Suman Ravuri and Shakir Mohamed and Mihaela Rosca and Oriol Vinyals},
      year={2018},
      eprint={1806.11006},
      archivePrefix={arXiv},
      primaryClass={cs.LG}
}

@inproceedings{bucilua2006model,
author = {Bucilua, Cristian and Caruana, Rich and Niculescu-Mizil, Alexandru},
title = {Model Compression},
year = {2006},
publisher = {Association for Computing Machinery},
address = {New York, NY, USA},
url = {https://doi.org/10.1145/1150402.1150464},
doi = {10.1145/1150402.1150464}
}

@misc{hinton2015distilling,
      title={Distilling the Knowledge in a Neural Network}, 
      author={Geoffrey Hinton and Oriol Vinyals and Jeff Dean},
      year={2015},
      eprint={1503.02531},
      archivePrefix={arXiv},
      primaryClass={stat.ML}
}

@article{jiao2019tinybert,
  title={Tinybert: Distilling bert for natural language understanding},
  author={Jiao, Xiaoqi and Yin, Yichun and Shang, Lifeng and Jiang, Xin and Chen, Xiao and Li, Linlin and Wang, Fang and Liu, Qun},
  journal={arXiv preprint arXiv:1909.10351},
  year={2019}
}

@inproceedings{chebotar2016distilling,
  title={Distilling Knowledge from Ensembles of Neural Networks for Speech Recognition.},
  author={Chebotar, Yevgen and Waters, Austin},
  booktitle={Interspeech},
  pages={3439--3443},
  year={2016}
}

@inproceedings{kim2019knowledge,
  title={Knowledge distillation using output errors for self-attention end-to-end models},
  author={Kim, Ho-Gyeong and Na, Hwidong and Lee, Hoshik and Lee, Jihyun and Kang, Tae Gyoon and Lee, Min-Joong and Choi, Young Sang},
  booktitle={ICASSP 2019-2019 IEEE International Conference on Acoustics, Speech and Signal Processing (ICASSP)},
  pages={6181--6185},
  year={2019},
  organization={IEEE}
}

@inproceedings{Fukuda2017EfficientKD,
  title={Efficient Knowledge Distillation from an Ensemble of Teachers},
  author={T. Fukuda and Masayuki Suzuki and Gakuto Kurata and S. Thomas and Jia Cui and B. Ramabhadran},
  booktitle={INTERSPEECH},
  year={2017}
}

@inproceedings{watanabe2017student,
  title={Student-teacher network learning with enhanced features},
  author={Watanabe, Shinji and Hori, Takaaki and Le Roux, Jonathan and Hershey, John R},
  booktitle={2017 IEEE International Conference on Acoustics, Speech and Signal Processing (ICASSP)},
  pages={5275--5279},
  year={2017},
  organization={IEEE}
}

@article{gao2020distilling,
  title={Distilling Knowledge from Ensembles of Acoustic Models for Joint CTC-Attention End-to-End Speech Recognition},
  author={Gao, Yan and Parcollet, Titouan and Lane, Nicholas},
  journal={arXiv preprint arXiv:2005.09310},
  year={2020}
}

@misc{turc2019wellread,
      title={Well-Read Students Learn Better: On the Importance of Pre-training Compact Models}, 
      author={Iulia Turc and Ming-Wei Chang and Kenton Lee and Kristina Toutanova},
      year={2019},
      eprint={1908.08962},
      archivePrefix={arXiv},
      primaryClass={cs.CL}
}

@article{sun2020contrastive,
  title={Contrastive Distillation on Intermediate Representations for Language Model Compression},
  author={Sun, Siqi and Gan, Zhe and Cheng, Yu and Fang, Yuwei and Wang, Shuohang and Liu, Jingjing},
  journal={arXiv preprint arXiv:2009.14167},
  year={2020}
}

@misc{sanh2020distilbert,
      title={DistilBERT, a distilled version of BERT: smaller, faster, cheaper and lighter}, 
      author={Victor Sanh and Lysandre Debut and Julien Chaumond and Thomas Wolf},
      year={2020},
      eprint={1910.01108},
      archivePrefix={arXiv},
      primaryClass={cs.CL}
}

@misc{furlanello2018born,
      title={Born Again Neural Networks}, 
      author={Tommaso Furlanello and Zachary C. Lipton and Michael Tschannen and Laurent Itti and Anima Anandkumar},
      year={2018},
      eprint={1805.04770},
      archivePrefix={arXiv},
      primaryClass={stat.ML}
}

@misc{sun2019patient,
      title={Patient Knowledge Distillation for BERT Model Compression}, 
      author={Siqi Sun and Yu Cheng and Zhe Gan and Jingjing Liu},
      year={2019},
      eprint={1908.09355},
      archivePrefix={arXiv},
      primaryClass={cs.CL}
}

@article{du2019implicit,
  title={Implicit generation and modeling with energy based models},
  author={Du, Yilun and Mordatch, Igor},
  journal={Advances in Neural Information Processing Systems},
  volume={32},
  pages={3608--3618},
  year={2019}
}

\appendix

\section{Samples}
\begin{figure}[H] 
\centering
\includegraphics[height=21cm, width=14cm]{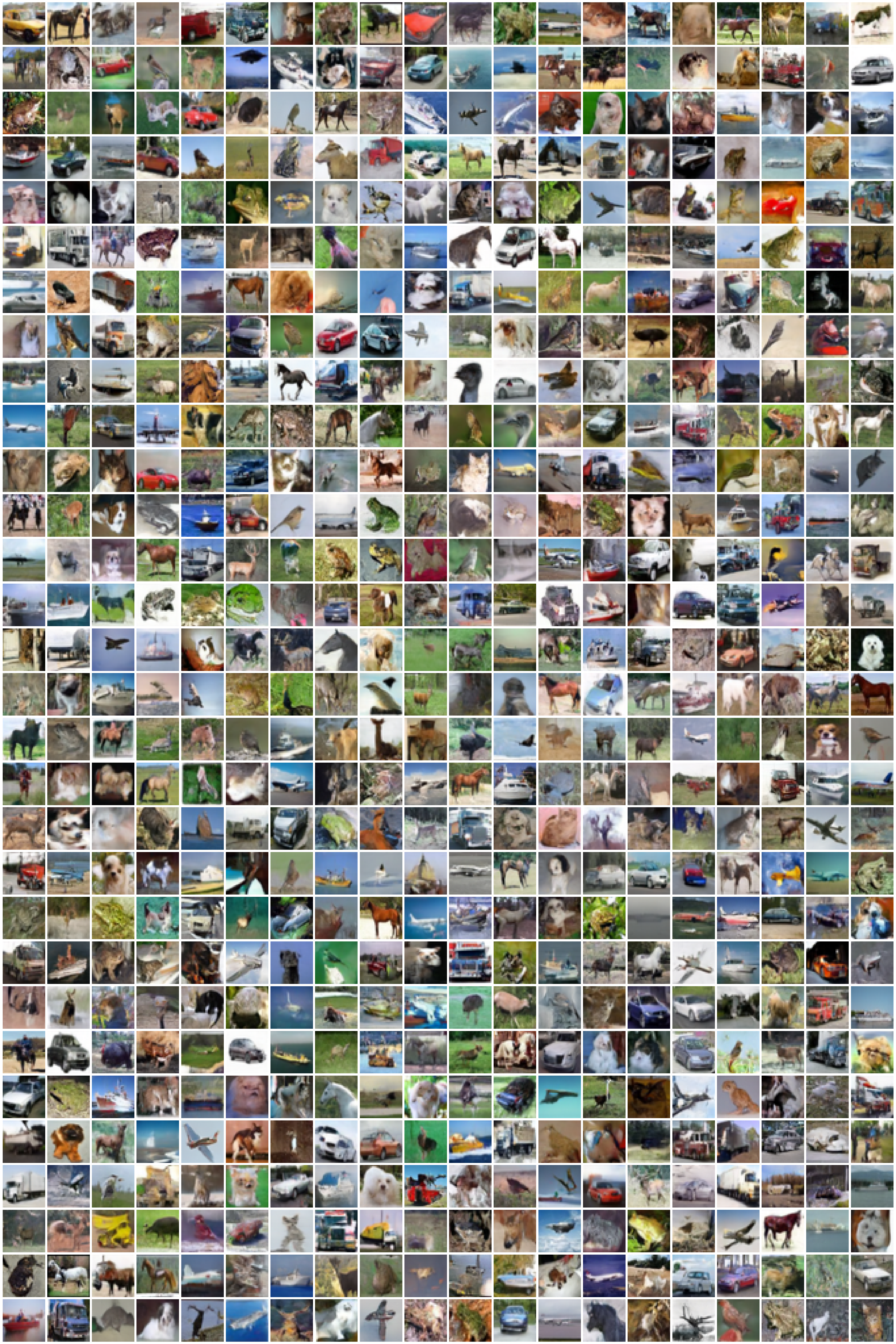}
\caption{Uncurated Samples from our CIFAR-10 model}
\label{fig6cifar_uncurated}
\end{figure}

\begin{figure}[H] 
\centering
\includegraphics[height=20cm, width=12cm]{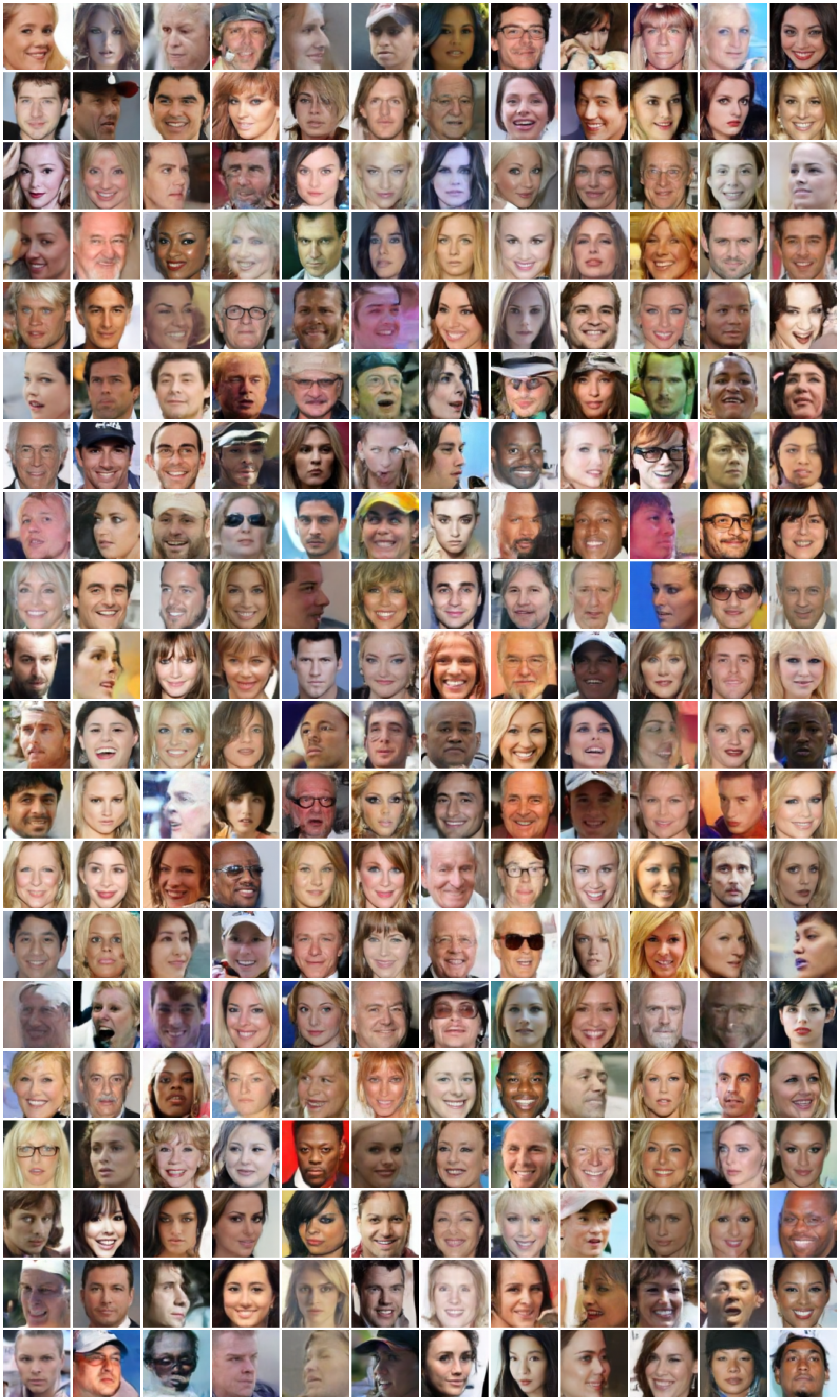}
\caption{Uncurated Samples from our CelebA model}
\label{fig7celeba_uncurated}
\end{figure}

\begin{figure}[H] 
\centering
\includegraphics[height=20.8cm, width=13cm]{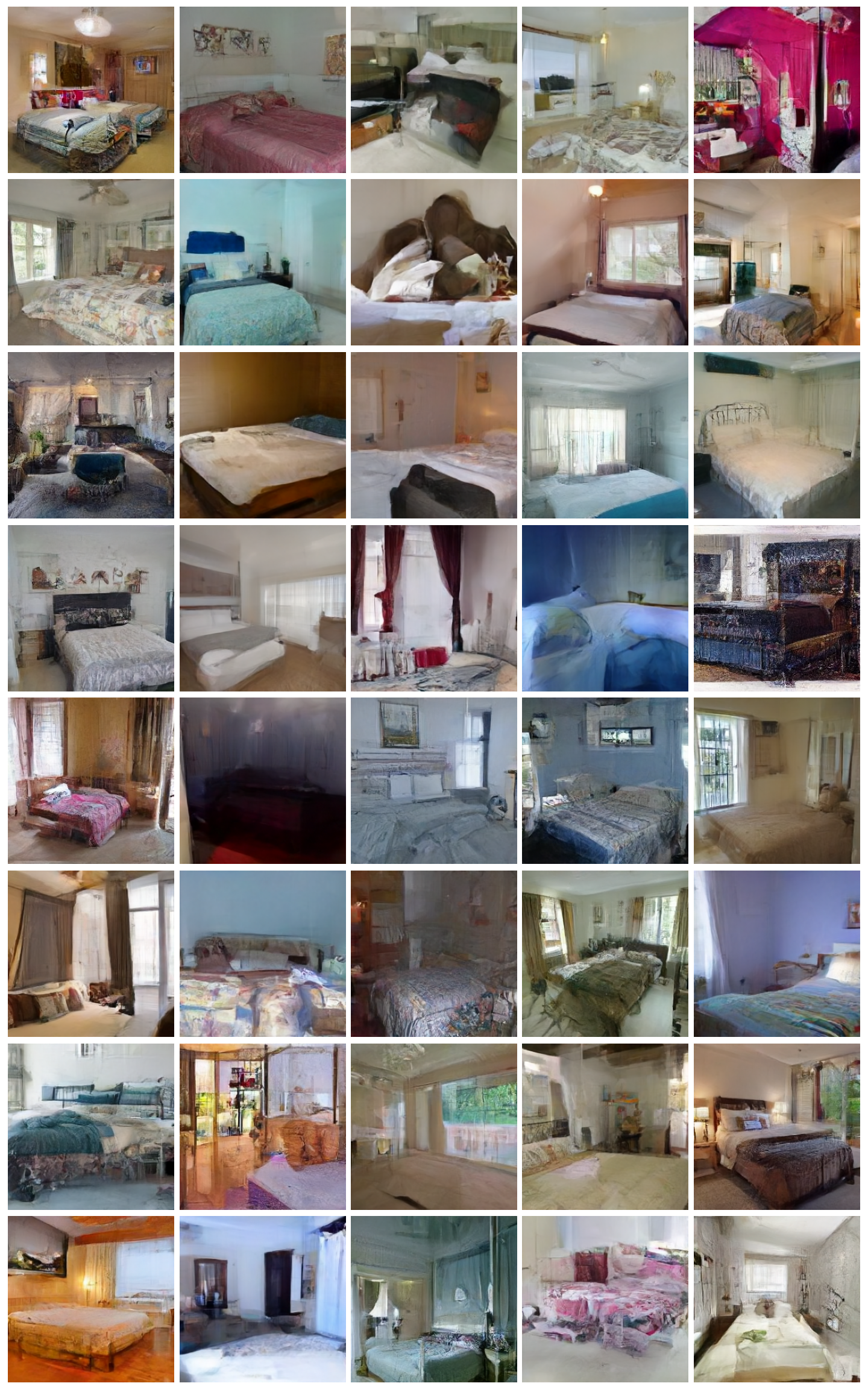}
\caption{Uncurated Samples from our LSUN Bedroom model}
\label{fig8lsun_bedroom_uncurated}
\end{figure}

\begin{figure}[H] 
\centering
\includegraphics[height=20.8cm, width=13cm]{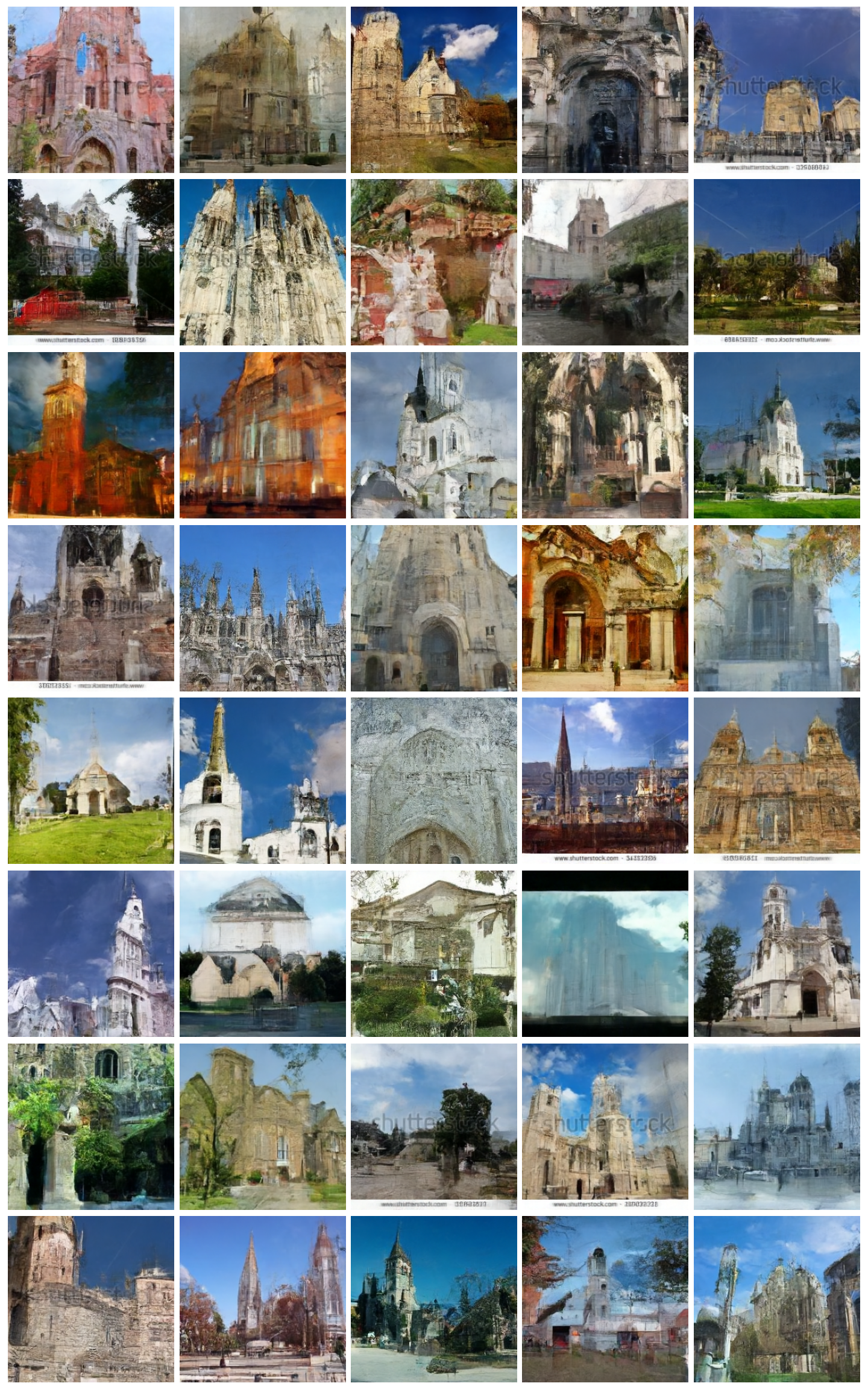}
\caption{Uncurated Samples from our LSUN Church-Outdoor model.}
\label{fig9lsun_church_uncurated}
\end{figure}

We show additional samples in Figure \ref{fig6cifar_uncurated} (CIFAR-10), \ref{fig7celeba_uncurated} (CelebA), \ref{fig8lsun_bedroom_uncurated} (Bedroom), and  \ref{fig9lsun_church_uncurated} (Church). Additional interpolation results can be seen in \ref{fig11interpolation_results}, and different size images can be seen in Figure \ref{fig12different_resolution}.

\begin{figure}[H] 
\centering
\includegraphics[height=8.4cm, width=16cm]{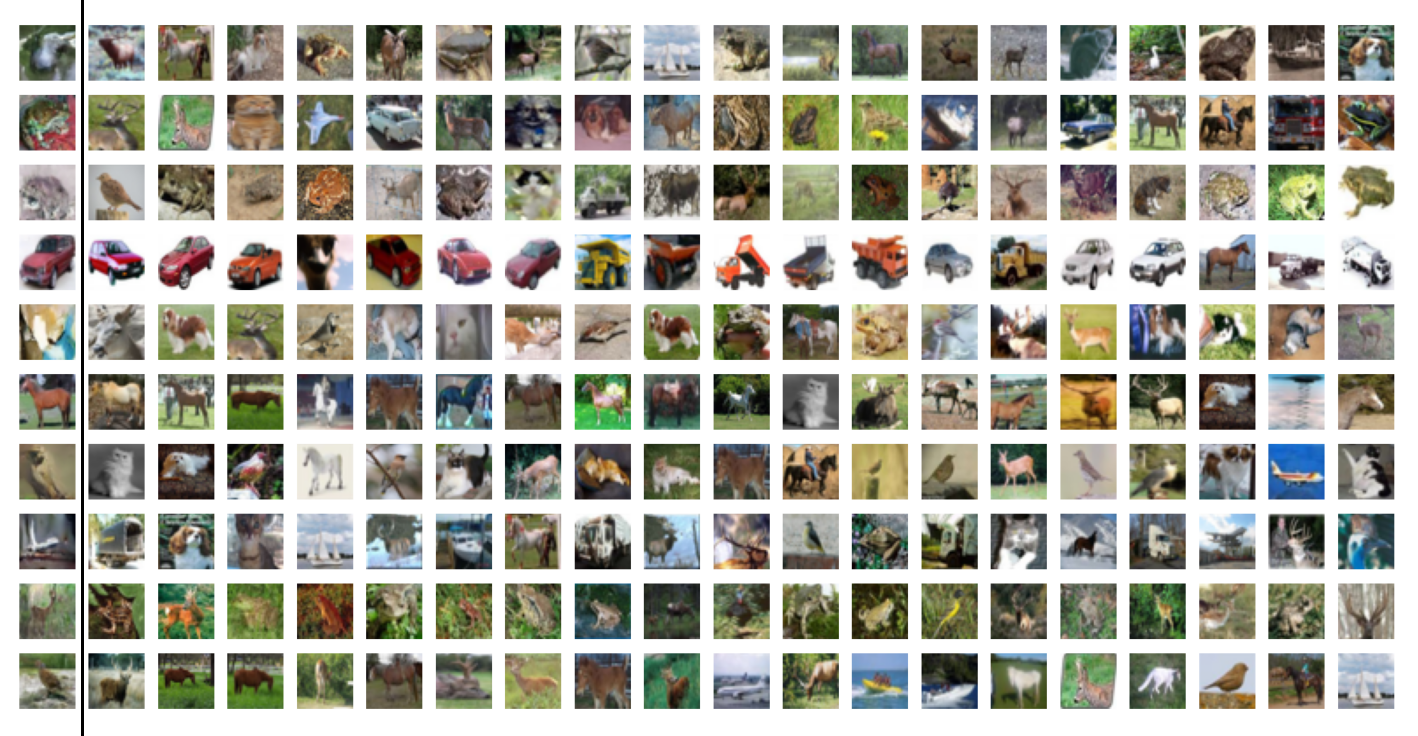}
\caption{Inception feature space nearest neighbors for CIFAR-10. Images generated from our model are in the leftmost column, and to the right of the black line are the nearest neighbors in the training set.}
\label{fig10inceptionspace_nearestneighbors}
\end{figure}

\begin{figure}[H] 
\centering
\includegraphics[height=9cm, width=16.5cm]{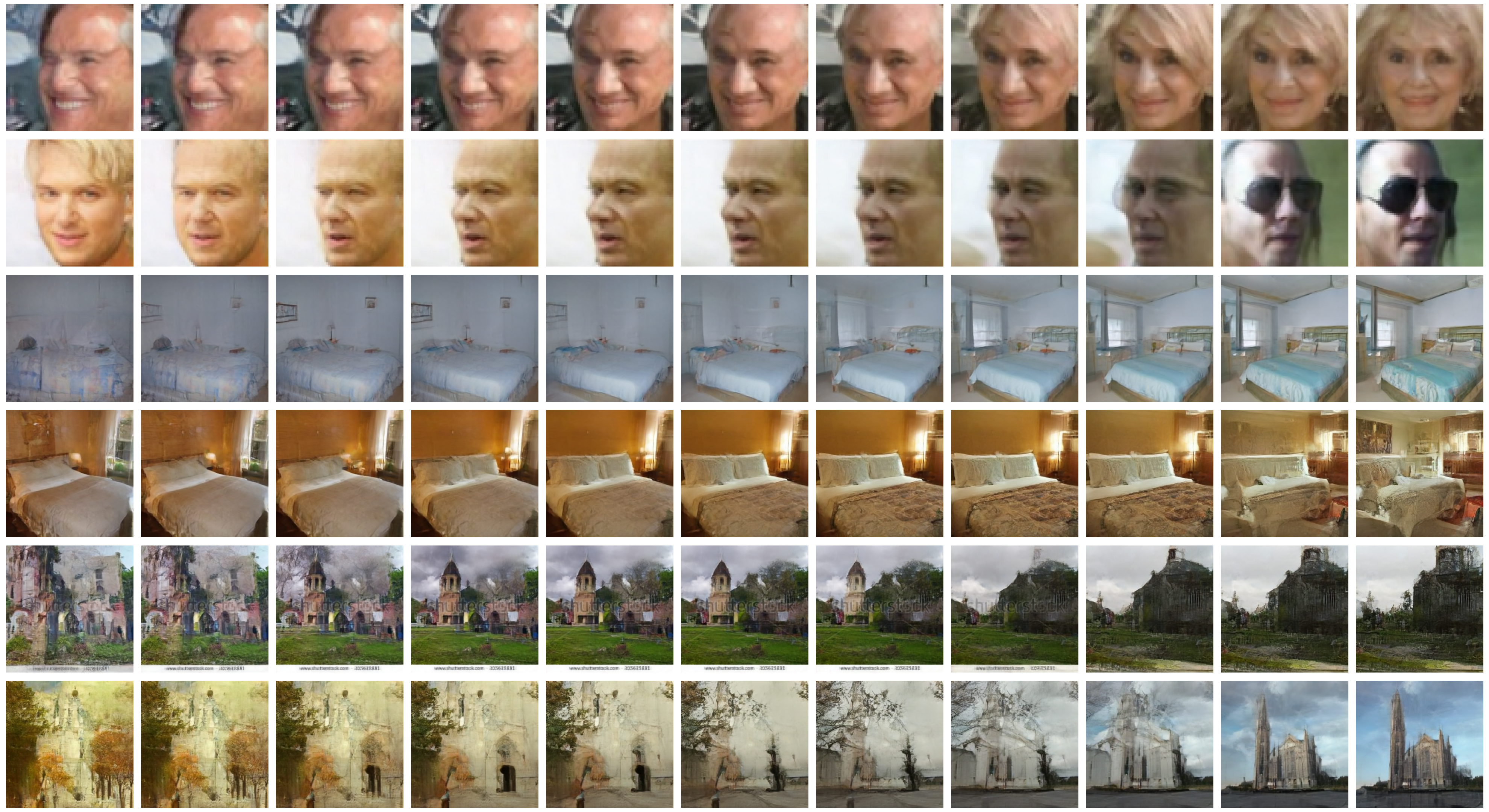}
\caption{Extended Interpolation results. Rows 1 and 2: CelebA images, Rows 3 and 4: LSUN Bedroom, Rows 5 and 6: LSUN church}
\label{fig11interpolation_results}
\end{figure}

\begin{figure}[H] 
\centering
\includegraphics[height=10.3cm, width=16.0cm]{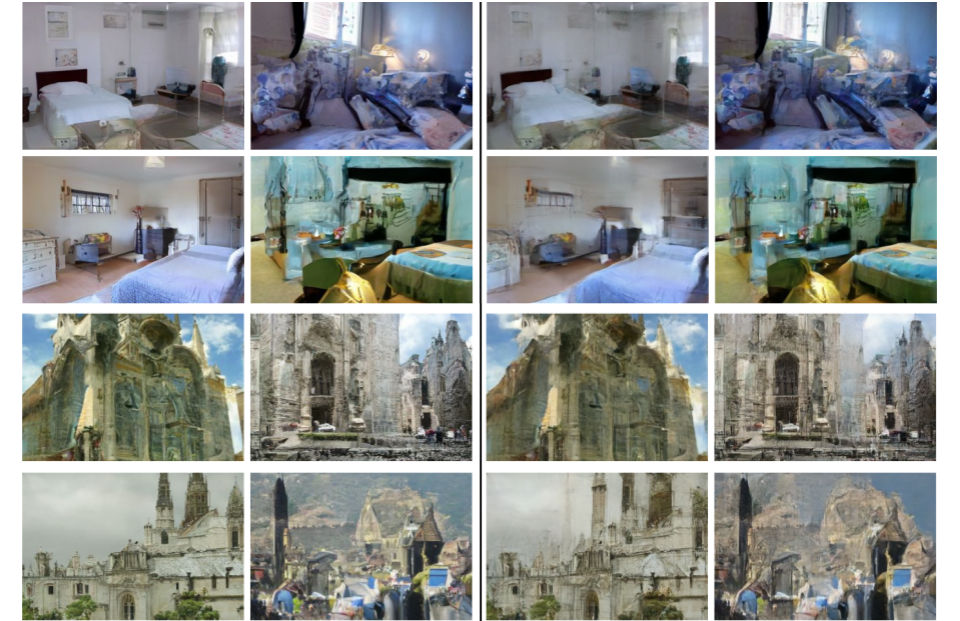}
\caption{Uncurated $256 \times 384$ images from a 50 step DDIM (teacher, left) and our model (right). Neither model has ever seen a $256 \times 384$ image in training, yet both produce decent samples, demonstrating that both the teacher and student generalize well.}
\label{fig12different_resolution}
\end{figure}

Since our model was not trained on real data, but instead on the teacher's output, we do not expect it to memorize training examples. To demonstrate this, we include nearest neighbor visualizations in Figure \ref{fig10inceptionspace_nearestneighbors}. 

\section{Experimental Details} \label{experimentaldetails}
A list of the hyperparameters we used can be seen in Table \ref{tab:hparams}. We used the Adam (\citep{adam}) optimizer. We linearly increased the learning rate for a number of warmup steps, and used no decay. We did not use dropout or any other regularization, and used test loss evaluated on unseen images for early stopping. We made some interesting observations when training, and list them as follows:
\begin{itemize} 
\item{For CIFAR-10, we found that training on L1 distance yielded better results than L2.}
\item{For LSUN Bedroom, we initially clipped the values of the gradients to $\pm 1.0$, but observed that the gradients were very large (norm consistently $> 10^5$). We found that \textit{not clipping} by value improved stability.} 
\item{For LSUN Bedroom, we initially chose $\beta_1 = 0.9, lr = 2 \times 10^{-5}$, but found that  \textit{increasing} $\beta_1$ to 0.98 and decreasing the learning rate to $5 \times 10^{-6}$ helped stabilize  training. We applied these to Church without sweeping.}
\end{itemize}

\begin{table}[h!]
  \begin{center}
    \caption{Hyperparameters used in our experiments.} 
    \label{tab:hparams}
    \begin{tabular}{l c c c c} 
      \\ 
      Model & CIFAR-10 & CelebA & LSUN Bedroom & LSUN Church\\
      \hline \\
      Parameters (M) & 35.7 & 78.7 & 114 & 114 \\
      Batch size & 512 & 512 & 32 & 32 \\
      Learning Rate & $2 \times 10^{-4}$ & $5 \times 10^{-5}$ & $5 \times 10^{-6}$ & $5 \times 10^{-6}$ \\
      Adam $\beta_1$ & 0.9 & 0.9 & 0.98 & 0.98 \\
      Adam $\beta_2$ & 0.98 & 0.98 & 0.999 & 0.999 \\
      EMA decay rate & 0.995 & 0.995 & 0.9995 & 0.9995 \\
      Warmup steps & 5000 & 5000 & 1000 & 1000 \\
      Max training iterations & 50k & 64k & 320k & 256k \\
      
    \end{tabular}
  \end{center}
\end{table}

\end{document}